\documentclass[10pt]{article}
\usepackage{url}
\usepackage{multirow}
\usepackage{amssymb}
\usepackage{graphicx}
\usepackage{color}
\usepackage{hyperref}

\usepackage[dvipsnames]{xcolor}

\DeclareGraphicsExtensions{.pdf,.png,.jpg}
\begin{document}
%

\newcommand{\mr}[1]{}
\newcommand{\eat}[1]{}

\newcommand{\term}[1]{\small {\tt #1}}
\newcommand{\triple}[1]{\small {$\langle${\tt #1}$\rangle$}}

\title{TeKnowbase: Towards Construction of a Knowledge-base for Technical Concepts}
\author{Prajna Upadhyay\\IIT Delhi \and Tanuma Patra\\IIT Delhi \and Ashwini Purkar\\IIT Delhi \and Maya Ramanath\\IIT Delhi}
\date{}

\maketitle
\begin{abstract}
In this paper, we describe the construction of TeKnowbase, a knowledge-base of technical concepts in computer science. Our main information sources are technical websites such as Webopedia and Techtarget as well as Wikipedia and online textbooks. We divide the knowledge-base construction problem into two parts -- the acquisition of entities and the extraction of relationships among these entities. Our knowledge-base consists of approximately 100,000 triples. We conducted an evaluation on a sample of triples and report an accuracy of a little over 90\%. We additionally conducted classification experiments on StackOverflow data with features from TeKnowbase and achieved improved classification accuracy. 
\end{abstract}

\section{Introduction}

As digital information gains more and more prominence, there is now a trend to organize this information to make it easier to query and to derive insights. One such trend is the creation of large \emph{knowledge-bases} (KBs) -- repositories of crisp, precise information, which are machine readable. Indeed the creation of such knowledge-bases has been a goal for decades now with projects such as Cyc \cite{Lenat:1995it} and Wordnet \cite{Miller:1995wk}.

With advances in information extraction research, and the availability of large amounts of structured and unstructured (textual) data, \emph{automatic} construction of knowledge-bases are not only possible, but also desirable because of the coverage they can offer. There are already many such general-purpose knowledge-bases such as Yago \cite{Suchanek:2007vb} and DBPedia \cite{Lehmann:2015wl}. Moreover, projects such as OpenIE \cite{Banko:2007ul} and NELL \cite{Carlson:2010vd} aim to extract information from unstructured textual sources on a large scale.

However, as the technology to automatically build large knowledge-bases matures, there is a paucity of high quality and \emph{specialized} KBs for specific domains. For some domains, for example, for the bio-medical domain, there are well-curated ontologies which partially address this gap (see, for example, the Gene Ontology project \cite{Ashburner:2000ja}). However, for domains such as Computer Science or IT in general, where such curation efforts are hard and the field itself is rapidly growing, it becomes critical to revisit the automatic construction processes that take advantage of domain-specific resources.

In this paper, our aim is to automatically construct a \emph{technical} knowledge-base, called TeKnowBase, of computer science concepts. One of the most important tasks in building any such "vertical" KB is the identification of the right kinds of resources. For example, even though Wikipedia contains technical content, identifying the right subset of this content is crucial. Similarly, while free online technical content, such as textbooks are available, it is important to identify what kind of extractions are possible. The identification of the right resources can sometimes yield bigger gains than using an elaborate information extraction technique.

A preliminary examination of computer science-related resources show that information can be extracted from many different kinds of sources, including, Wikipedia, technical websites such as Webopedia, online textbooks, technical question answer fora such as StackOverflow, etc. By studying these resources more closely, we developed simple, but effective techniques to build TeKnowbase. Our first step is to acquire a dictionary of concepts and entities relevant to computer science. Using this dictionary, we can further extract relationships among them. We make use of the semantic web standard, RDF, where information is represented as triples of the form $\langle$subject$\rangle$$\langle$predicate$\rangle$$\langle$object$\rangle$ -- in a nutshell, each triple makes a statement about the $\langle$subject$\rangle$. Table \ref{tab:examples} shows examples of the kind of triples we extract and the number of such triples in our knowledge-base.

\begin{table*}
\centering
\begin{tabular}{|l|l|l|}
\hline
Relation & Example & \# Triples \\ 
\hline
\term{type} & \term{Topological\_sorting typeOf Graph\_algorithms} & 44,221  \\ 
\hline
\term{concept} & \triple{Nash\_equilibrium conceptsOf Game\_theory} & 833 \\
\hline
\term{subTopic} &\triple{Hamming\_code subTopicOf Algebraic\_Coding\_Theory} & 1520 \\
\hline
\term{application} & \triple{Group\_testing applicationOf Coding\_theory} & 1650 \\ 
\hline
\term{terminology} & \triple{Blob\_detection isTerminology Image\_Processing} & 32,722 \\  
\hline
\end{tabular}
\caption{Statistics for and examples of a subset of relationships}
\label{tab:examples}
\end{table*}

Such a knowledge-base, that organizes the space of technical concepts in a systematic way, can be used in a variety of applications. For example, as with general-purpose knowledge-bases, a technical KB can be used for improving classification accuracy, disambiguation of text, linking entities, and, in general, semantic search. Moreover, a technical KB can be of use in a variety of learning scenarios. For example, to students wanting to learn about a new concept, a technical KB can be a valuable resource in identifying related concepts and perhaps pre-requisites. Another useful application is the automatic generation of questions to test general, and conceptual knowledge. For example, a question such as "Name some applications of coding theory" can be generated and the answer graded automatically (see a similar example in \cite{Seyler:2015we} in the context of general-purpose KBs).

Note that the set of technical concepts that we consider here are limited to \emph{named} entities. In addition to these, there could be many unnamed entities such as theorem statements, formulae, equations, algorithm listings, etc.\footnote{Note that, named entities, such as \term{heap sort}, even though an algorithm are still included in our KB.} In principle, we will be able to recognize these entities and provide a system-generated name to them. But, it is challenging to go beyond this to provide means of querying these entities. In this paper, we limit our scope to only named entities.

\paragraph*{Contributions}
Our contributions are as follows:
\begin{itemize}
	\item We describe the construction of TeKnowbase, a knowledge-base of technical concepts in computer science. Our approach is general enough that it can be used for other subjects as well.
	\item In order to construct TeKnowbase, we carefully studied different resources which could be helpful and systematically explored information extraction techniques to collect triples for our KB. We highlight their strengths and drawbacks.
	\item We present a study on the quality of triples in TeKnowbase that shows a precision of about 90\%.
	\item We perform a simple classification experiment using features from TeKnowbase to improve the classification accuracy of technical posts on StackOverflow, a technical question-answer forum.
\end{itemize}

\paragraph*{Organization} We briefly describe related work in Section \ref{sec:rel-work}. Our main contributions are described in Sections \ref{sec:concepts} and \ref{sec:relations}. We present an evaluation of our knowledge-base in Section \ref{sec:evaluation}. We conclude  and identify avenues for future work in Section \ref{sec:conc}.

\section{Related Work}
\label{sec:rel-work}

Knowledge-bases have many applications -- see for example, Google's Knowledge Graph \cite{googleKG} which is used to provide users with concise summaries to queries about entities, semantic search engines \cite{Tummarello:2010te}, question answering systems such as IBM Watson \cite{Ferrucci:2010vl}, etc. Our aim is to build a domain-specific knowledge-base (in this case, computer science) which helps in these kinds of applications, but in addition can serve as a resource for learning, including, for example, generating a reading order of concepts in order to learn a new one, automatically generating test or quiz questions about concepts to test student understanding, etc.

Recently, systems which facilitate knowledge-base construction from heterogeneous sources have been proposed. For example, DeepDive \cite{DeSa:2016dx} aims to consume a large number of heterogeneous data sources for extraction and combines evidence from different sources to output a probabilistic KB. Similarly, Google's Knowledge Vault \cite{Dong:2014fp} also aims to fuse data from multiple resources on the Web to construct a KB. Our effort is similar in that we make use of heterogeneous data sources and customise our extractions. However, since our focus is quite narrow and we use very few sources, we do not perform any inferencing.

\paragraph*{Entity extraction}
One of the important aspects of building domain-specific knowledge-bases is that a dictionary of terms that are relevant to the domain should be acquired. It is possible that such dictionaries are already available (for example, lists of people), but for others, we need techniques to build this dictionary. \cite{Ren:2016um} gives an overview of supervised and unsupervised methods to recognize entities from text. We follow a more straightforward approach -- we specifically target technology websites and write wrappers to extract a list of entities related to computer science.

\paragraph*{Information Extraction}
Research in information extraction to build knowledge-bases make use of a variety of techniques (see \cite{Suchanek:2014tt} for an overview). In general, information extraction can be done from mostly structured resources such as Wikipedia (see, for example, YAGO \cite{Suchanek:2007vb}) or from unstructured sources (for example, OpenIE \cite{Banko:2007ul}) where the relations are not known ahead of time. Moreover, there are rule-based systems such as SystemT \cite{Li:2011un}, using surface patterns and supervised techniques for known relations, distant supervision, etc. (see, for example, Hearst patterns \cite{Hearst:1992uz} and \cite{Mintz:2009tn}, \cite{Carlson:2010va}). We use a mix of these approaches -- we formulate different ways to exploit the structured information sources in Wikipedia, and use surface patterns to extract relationships from unstructured sources, such as online books. Some of these techniques provided us with high quality triples, while others failed. We analyze both our successes as well as failures in the paper.
\section{TeKnowBase: Acquiring a list of concepts}
\label{sec:concepts}

Our strategy to construct TeKnowbase was to first construct a dictionary of technical concepts and subsequently, to use this dictionary to annotate text and acquire triples. We found that various popular named entity taggers were inadequate in our domain. Therefore, we decided to build our own dictionary from various resources. We converged on three kinds of resources -- the oft-used (semi-)structured resource Wikipedia, technology-specific websites Webopedia\footnote{\url{http://www.webopedia.com}} and TechTarget\footnote{\url{http://www.techtarget.com/}}, and subject-specific online textbooks.

Our idea was to use well-structured information from these resources to construct the dictionary of entities. Subsequently, we studied the various structured pieces of information in Wikipedia to extract triples by appropriately engineering our code, and used well-known IE techniques to extract triples from unstructured texts. We describe our ideas in the rest of this section.

\subsection{Building a dictionary of technical concepts}
There are two main challenges in building a list of technical concepts: i) finding the right resources to extract the concepts (which we previously mentioned), and, ii) reconciling different forms of the same entities (also called the entity resolution problem).

We made use of Wikipedia, technical websites, and online textbooks to build our dictionary. We used Wikipedia's article titles as well as it's category system as a source of concepts. Our corpus of Wikipedia articles consists of all articles under the super category \term{Computing}. In all, there were approximately 60,000 articles. The titles of each article was considered an entity. Examples entities we found were \term{Heap\_Sort}, \term{Naive-Bayes\_Classifier}, etc.

Our second set of resources were two websites Webopedia and TechTarget. Each website consists of a number technical terms and their definitions in a specific format. From both these websites, we extracted approximately 26,500 entities.

Finally, we extracted entities from the indexes of 6 online textbooks (indexes are also well-structured). These textbooks were specific to the IR and ML domains. The idea can be extended to multiple such online textbooks that are freely available. In all, we extracted approximately 16,500 entities from these textbooks.

While Wikipedia has articles on a number of technical concepts, it is not exhaustive. For example, the terms \term{average\_page\_depth} (related to Web Analytics) and \term{Herbrand\_universe} (related to logic) could not be found in Wikipedia, but were found on the technical websites and textbooks respectively.

\subsection{Resolving entities}
\label{sec:er}
Clearly, the lists of raw concepts extracted from each source have overlaps. We used edit distance to identify and remove duplicates. However, we found that edit distance by itself was not sufficient to resolve all entities, because there are numerous acronyms and abbreviations that are commonly used. Since we wanted to retain both the acronym as well as its expansion as separate entities, we treated the problem of finding (acronym, expansion) pairs as a triple extraction problem -- specifically triples for the \term{synonymOf} relation. Since this involves triple extraction from unstructured sources, we defer the description of our technique to Section \ref{sec:relations}. 

In all, our entity list consists of 85,000 entities after entity resolution.

\section{Acquiring relationships between concepts}
\label{sec:relations}
We classified our relationship extraction task into four types, based on the sources (structured and unstructured) and our prior knowledge of the relations (known relations and unknown relations). By studying our sources and our own idea of the kinds of relations that we wanted in our KB, we made a list of relations to extract (this was our set of known relations). Most importantly, we wanted to construct a taxonomy of concepts, and therefore, the \term{typeOf} relation was included. It was obvious that we would be unable to list all possible relations, and therefore, we made use of several techniques to acquire unknown relations. We give a brief overview below and describe our techniques in detail from Section \ref{sec:sk}.

\begin{itemize}
	\item {\bf Structured source, known relations}: As mentioned previously, we were able to recognize, by manual inspection, that the source already contains some of the relationships we want to extract, but organized in a different way. We need to convert one kind of structure (available from the source), into another kind of structure (RDF triples). Wikipedia contains such structured information which can be exploited -- specifically, certain types of pages and the organization of the content in a page.
	\item {\bf Structured source, unknown relations}: Wikipedia also provides us with other kinds of structured information such as template information -- that is, tabular data with row headers. These headers potentially describe relationships, but it is not possible to know ahead of time exactly what these relationships are.
	\item {\bf Unstructured source, known relations}: Unstructured sources are textual sources, such as online textbooks. Our list of entities, constructed as described in Section \ref{sec:concepts}, are of importance here. We could annotate entities in the textual sources and be confident of the correctness of these entities. Subsequently, we find relationships between these entities. We tried the simple technique of using surface patterns for our relations. As we report later in the paper, we were only partially successful in our efforts.	
	\item {\bf Unstructured source, unknown relations}: Finally, we annotated the entities in the textual sources, and ran the open information extraction systems OLLIE \cite{Mausam:2012wu} to extract \emph{any} relationship between the entities.
\end{itemize}

\subsection{Structured source, known relation}
\label{sec:sk}
Since our aim was to construct a technical knowledge-base, we manually made a list of relations that our knowledge-base should contain -- this is our list of known relations. The relations included the taxonomic relation  \term{typeOf} (as in, \triple{JPEG typeOf file\_format}) and other interesting relationships such as \term{algorithmFor}, \term{subTopicOf}, \term{applicationOf}, \term{techniqueFor}, etc. In all, we identified 18 relationships that we felt were interesting and formulated techniques to extract them from Wikipedia. 

\begin{figure}[htbp]
\begin{minipage}{.5\textwidth}	
\centering
\includegraphics[scale=0.5]{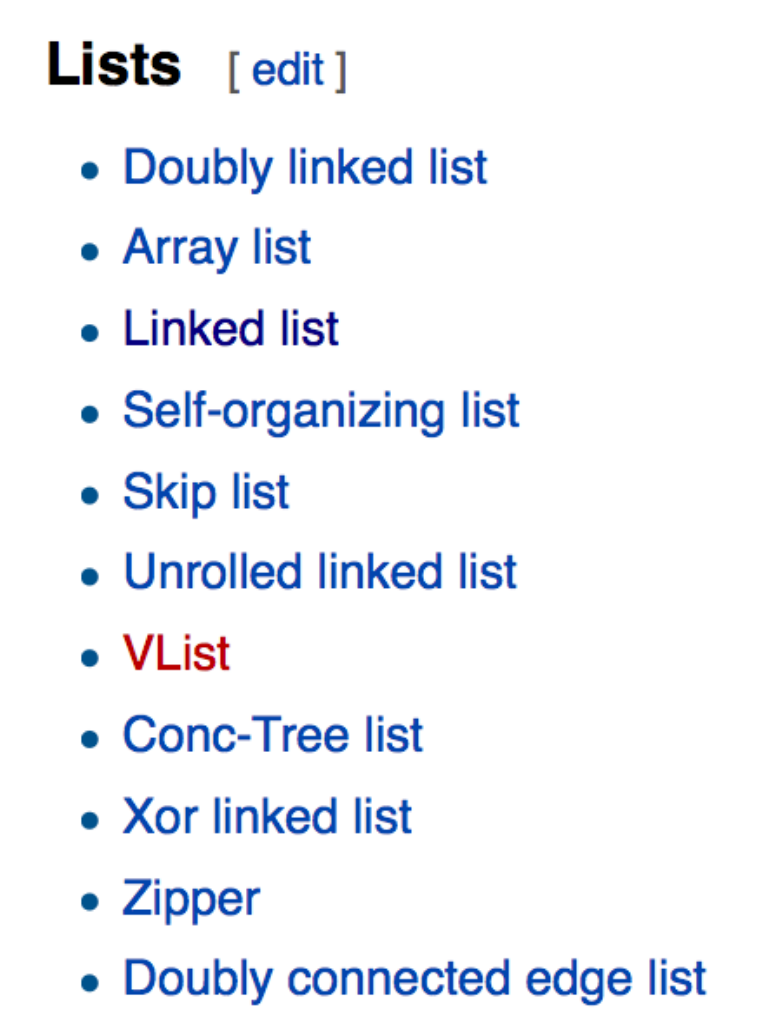}
\caption{Snippet from "List of Data Structures" page. Extraction of \term{typeOf} relations from the list structure.}
\label{fig:ds}
\end{minipage}
\begin{minipage}{.5\textwidth}	
\centering
\includegraphics[scale=0.5]{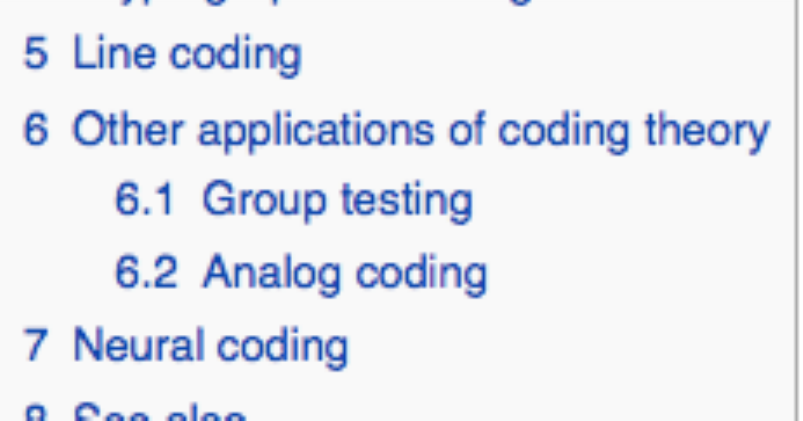}
\caption{Snippet of the TOC in "Coding theory" page. Extraction of \term{applicationOf} relation from the list/sublist structure.}
\label{fig:coding}
\end{minipage}
\end{figure}

\paragraph*{Overview pages.} We made use of two kinds of structured pages -- "List" pages and "Outline" pages (for example, the pages, \term{List of machine learning concepts}, \term{Outline of Cryptography}, etc.). These pages organize lists of entities with headings and sub-headings. Extracting this information gives us the relations \term{typeOf} and \term{subTopicOf} with good accuracy (see Section \ref{sec:evaluation} for an evaluation of these relations). Figure \ref{fig:ds} shows an example of a list page for data structures. We see a list of terms under a heading and can extract triples of the form \triple{xor\_linked\_list typeof list}. Further we were able to extract \emph{taxonomic hierarchies} of two levels by relating the headings to the article title. Continuing the previous example, \triple{List typeOf Data\_Structure} was extracted based on the article title.

\paragraph*{Articles on specific topics.} These pages refer to discussion on specific topics such as, say, 'Coding Theory'. These pages consist of many structured pieces of information. We made use of three of them of them as follows: 
	\begin{itemize}
		\item {\bf The table of contents (TOC)}: From our list of known relations, we searched for keywords within the TOC. If the keyword occurred in an item of the TOC, then the sub-items were likely to be related to it. For example, in Figure \ref{fig:coding}, the Coding Theory page consists of the following item in its TOC: 'Other applications of coding theory' and this in turn consists of two sub-items 'Group testing' and 'Analog coding'. Since one of the keywords from our known relations is 'application', and the page under consideration is Coding Theory, we extract the triples \triple{Group\_testing applicationOf Coding\_theory} and \triple{Analog\_coding applicationOf Coding\_theory}.
		\item {\bf Section-List within articles}: Next, there are several sub-headings in articles which consist of links to other topics. For example, the page on 'Automated Theorem Proving' consists of a subheading 'Popular Techniques' -- this section simply consists of a list of techniques which are linked to their wikipedia page. Since 'technique' is a keyword from our list of known relations, we identify this section-list pattern and acquire triples such as \triple{Model\_checking techniqueFor Automated\_Theorem\_Proving}.
		\item {\bf List hierarchies in articles}: As in the case of "List" pages and "Outline" pages, we make use of list hierarchies in articles to extract the \term{typeOf} relationships.
	\end{itemize}

Table \ref{tab:wikisources} gives a summary of the number of triples we extracted from each of these sources.
\begin{table}
\centering
\begin{tabular}{l|l|l}
{\bf Source} & {\bf \# found} & {\bf \# triples}\\
\hline
List, Outline pages& 503 & 35835\\
TOC & 1838 & 7412\\
Section-List & 1909 & 10191\\
List hierarchies & 113 & 12679\\
Templates & 1139 & 30434\\
\hline
\end{tabular}
\caption{Wikipedia sources from which triples were extracted}
\label{tab:wikisources}
\end{table}

\begin{figure}[ht]
\includegraphics[scale=0.5]{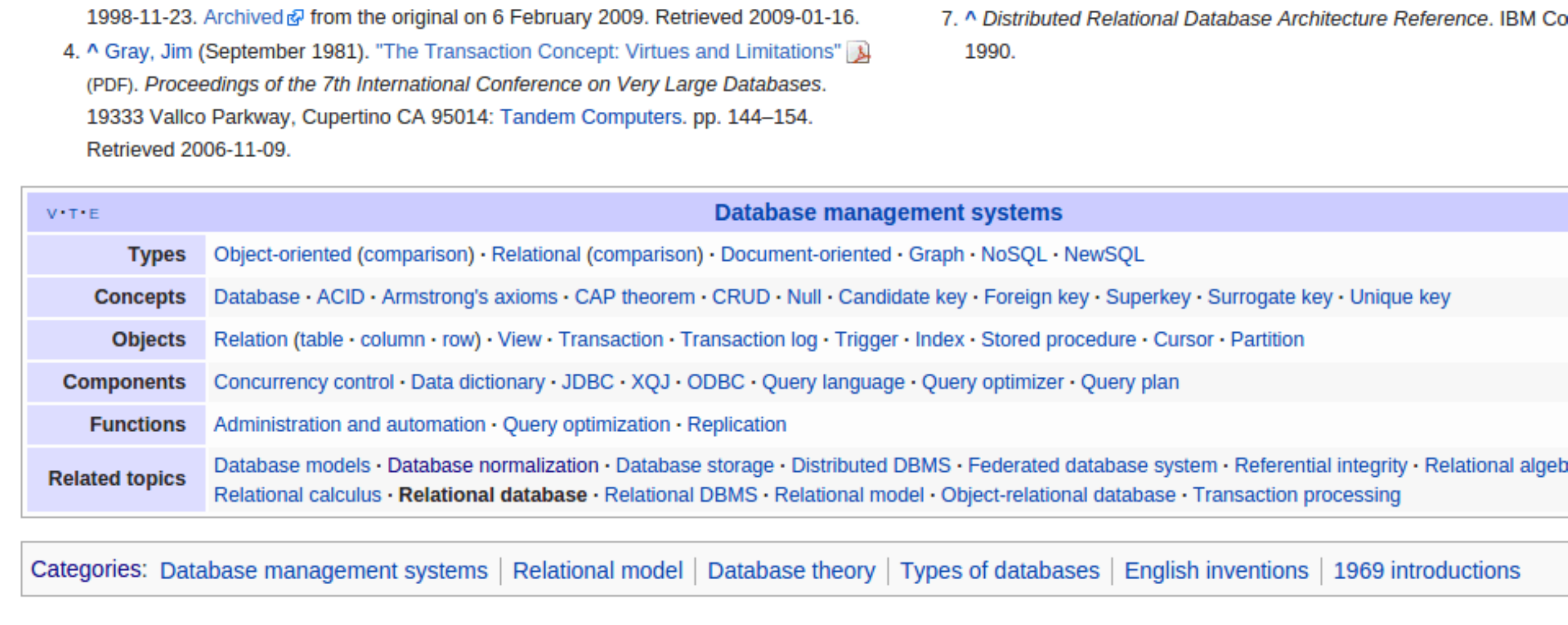}
\caption{DBMS Template}
\label{fig:dbms-template}
\end{figure}

\subsection{Structured source, unknown relations}
Continuing with Wikipedia as our nearly structured source, we next set about extracting relationships that were not in our list. For this purpose, we targeted template information that are available in many articles. For example, Figure \ref{fig:dbms-template} shows information about Database Management Systems. We made use of the row headers as new relations. For example, in the figure, we have row headings like 'Concepts', 'Objects', 'Functions', etc. which can serve as new relations \term{conceptOf}, 'functionOf', etc, leading to triples such as \triple{Query\_Optimization functionOf Database\_Management\_Systems}.

\subsection{Unstructured sources}
\label{sec:unstructured}
Our unstructured sources include textual description of terms in both Webopedia and Techtarget as well 6 online textbooks related to IR and ML. As previously mentioned, we first annotated the text from all these sources with entities from our entity dictionary. We then tried to extract relationships from them as follows:

\paragraph*{Known relations.} We formulated simple textual patterns for each of our known relations. For example, for the relation \term{typeOf}, we used the pattern "is a type of" and for the relation \term{algorithmFor}, we used the pattern "is an algorithm for". Our search for new triples simply found these patterns in the text and if there were annotated entities around the pattern, then these entities were taken as arguments for the relation. We were successful in identifying the \term{synonymOf} relation\footnote{Recall from Section \ref{sec:er} that we had both abbreviations as well as expansions in our dictionary and wanted to resolve them through identifying the \term{synonymOf} relation between them.} by using the patterns "is abbreviation for", "X (Y)" and "is short for" and extracted a little more than 1,000 such triples. Examples include terms like \term{JPEG}, which resolved to \term{Joint Photographic Experts Group}

	However, similar to the common experience with other types of corpora, surface patterns alone were not sufficient to extract accurate triples in technical corpora either. Since these extractions were not of high quality, we discarded them from the KB. Improving extractions are part of our future work.

\paragraph*{Unknown relations.} As our last method of extracting relationships, we made use of the open information extraction tool, OLLIE \cite{Mausam:2012wu}. Given our annotated textual corpus, we ran OLLIE to find any kind of relationship between entities. We were not very successful in extracting crisp relationships and we need to further study if and how open IE techniques can help us. 

Overall, TeKnowbase consists of nearly 100,000 triples. Some basic statistics are shown in Table \ref{tab:stats}. Examples of relation types and statistics of a few selected relations are given in Table \ref{tab:examples}

\begin{table}[h]
\centering
\begin{tabular}{ll}
\hline
No. of unique entities & 85162 \\
No. of unique relations &  1326 \\
Most frequent relation & \term{typeOf} with 44,221\\
Total no. of triples & 98,464  \\
No. of triples extracted from Wikipedia & 97,323 \\
No. of triples extracted from unstructured sources & 1141 \\
\hline
\end{tabular}
\caption{TeKnowbase Statistics} 
\label{tab:stats}
\end{table}

\section{Evaluation}
\label{sec:evaluation}

We performed two kinds of evaluations on TeKnowbase. First, we performed a direct evaluation on the quality of the extractions. We sampled a subset of triples and performed a user evaluation on the accuracy of these triples. Second, we used TeKnowbase in a classification experiment to see if features from this KB could improve the classification accuracy in the style of \cite{Gabrilovich:2005ta}. We describe each of these experiments below and report our results.

\subsection{Experiment 1: Evaluation of Quality}

\paragraph*{Setup.} We chose the top-5 most frequent relations for evaluation. These were: \term{typeOf}, \term{terminology}, \term{synonymOf} \term{subTopicOf} and \term{applicationOf}. Together, these five relations constitute about 84\% of the triples in our KB. We used stratified sampling to sample from each type of resource. Overall, we sampled 2\% of triples corresponding to each relation.

\paragraph*{Metrics.} Since there is no ground truth against which to evaluate these triples, we relied on user judgement. We asked graduate students to evaluate the triples with the help of the Wikipedia sources if required. Each triple was evaluated by two evaluators and we marked a triple as correct only if both evaluators agreed. 

\begin{table}
\centering
\begin{tabular}{|l|p{0.7in}|l|}
\hline
{\bf Relation} & {\bf \# Evaluated triples} & {\bf Accuracy}\\
\hline
\term{typeOf} & 851 & $82.5\% \pm 1.2\%$\\
\term{terminologyOf} & 606 & $93\% \pm 3.3\%$\\
\term{synonymOf} & 70 & $98\% \pm 0$\\
\term{subTopicOf} & 55 & $93.5\% \pm 2.5\%$\\
\term{applicationOf} & 40 & $89.7\% \pm 3.8\%$\\
\hline
\end{tabular}
\caption{Evaluation of a subset of triples in TeKnowbase.}	
\label{tab:accuracy}
\end{table}

\subsubsection{Results and Analysis} Table \ref{tab:accuracy} shows the accuracy of triples for each area. We computed the Wilson interval at $95\%$ confidence for each relation.

\paragraph*{The best and worst.}
On closer examination of these results, we found that we achieved the best results for the \term{synonym} relation. These triples consisted of both expansions of abbreviations, such as \term{ALU} and \term{Arithmetic Logic Unit} as well as alternate terminology such as \term{Photoshop} and \term{Adobe Photoshop}.

The best source of extractions are the Wikipedia list pages\footnote{Recall that Wikipedia list pages are those which list concepts and typically have an article title starting from \term{List\_of\_}}. In our list of top-5 relations, only 3 were extracted from Wikipedia list pages -- \term{typeOf}, \term{subtopicOf} and \term{synonymOf} -- and all of them were nearly $100\%$ accurate.

The major source of errors was the \term{typeOf} extractions from sources other than the list pages above and accounted for nearly 50\% of the errors in our evaluation set. Triples extracted from TOCs and Section-lists accounted for many of these errors. Recall from Section \ref{sec:sk} that we search for a keyword from our list of relations in the TOC items and associate the sublist of that item with the relation corresponding to that keyword. This heuristic did not always work well to identify the correct relation. For example, one of the errors was made when "Game types" was an item in the TOC of the page "Game Theory". It listed "Symmetric/Asymmetric" as a type of game, but we extracted \triple{Symmetric/Asymmetric typeOf Game\_theory} which is incorrect.

\paragraph*{Taxonomy.} We specifically analysed the taxonomy (all triples consisting of the \term{typeOf} relation) since this is an important subset of any KB as well as the largest subset in TeKnowbase. As previously mentioned, our taxonomy consists of over $44,000$ triples and our evaluation yielded an accuracy of $82.5\% \pm 1.2\%$ at $95\%$ confidence. The top two sources of these triples were: Wikipedia lists, List Hierarchies. Around 2000 distinct classes were identified, including, \term{file formats} (nearly 800 triples), \term{programming languages} (nearly 700 triples), etc. The accuracy of the \term{typeOf} triples is affected by different kinds of incorrect extractions from these two sources, even though they account for a very small percentage of the total errors. For example, the list of computer scientists is organized alphabetically and we failed to identify the correct class. This is because one of our heuristics is to use the section header as the class in the "List\_of" pages (refer to Figure \ref{fig:ds} in Section \ref{sec:sk}). 

Other triples were interesting, but not very useful. For example, we have several triples of type \term{TCP and UDP ports}. While these port numbers by themselves indeed belong to the class \term{TCP and UDP ports}, they are still just numbers with no indication of what services they are typically used for. 

Finally, our taxonomy consists of a limited hierarchy -- for example, \term{XOR linked list}, \term{List} and \term{Linear Data Structure} form a hierarchy of length two -- however, there are more opportunities to complete this hierarchy. For example, there are several \term{audio programming languages}, \term{C-family programming languages}, etc. which are in turn \term{programming languages}, but are currently not identified as such by our system. This is an important challenge for future work.

\subsection{Experiment 2: Classification}
In \cite{Gabrilovich:2005ta}, the authors showed how classification accuracy could be improved by generating features from domain specific, ontological knowledge. For example, a document belonging to the class "databases" may not actually contain the term "database", but simply have terms related to databases. If this relationship is explicitly captured in TeKnowbase, then that is a useful feature to add. We conducted a simple experiment to evaluate the use of TeKnowbase in a classification task.

\paragraph*{Setup.} StackOverflow\footnote{\url{stackoverflow.com}} is a forum for technical discussions. A page in the website consists of a question asked by a user followed by several answers to that question. The question itself may be tagged by the user with several hashtags. The administrators of the site classify the question into one of several technical categories. Our task is to classify a given question \emph{automatically} into a specific technical category.

We downloaded the StackOverflow data dump and chose questions from 3 different categories: "databases", "networking", and, "data-structures" We created a corpus of 1500  questions including the title (500 for each category). The category into which the questions were manually classified by the StackOverflow site were taken as the ground truth.

\paragraph*{Features generation.} First, we annotated the posts with entities from our dictionary. Entities as a whole were treated as a feature and were not broken up into separate words. We generated the following set of features for training.
\begin{itemize}
	\item {\bf BOW}: Bag of words model (note that entities remained whole).
	\item {\bf BOW-TKB}: In addition to the words and entities above, for each \emph{entity}, we generated an additional set of features by looking at the relationships the entities participated in. In particular, we used the following relations: \term{algorithm}, \term{definition}, \term{concept}, \term{topic}, \term{approach} and \term{method}. For example, if the entity \term{run length encoding} occurred in the post, then we added as a feature \term{data compression} since we have the triple \triple{run\_length\_encoding methodOf data\_compression}.
\end{itemize}

\paragraph*{Classification algorithms.} We trained both a Naive-bayes classifier as well as SVM with each of the feature sets above.

\paragraph*{Results.} We performed 5-fold cross validation with each classifier and feature set and report the accuracies in Table \ref{tab:classifier}. Clearly, simply adding new features from TeKnowbase helps in improving the accuracies of the classifiers. This result is encouraging and we expect that optimizing the addition of features (for example, coming up with heuristics to decide which relations to use) will result in further gains.

\begin{table}[ht]
\centering
\begin{tabular}{|c|c|c|c|}
\hline
&{\bf BOW} & {\bf BOW-TKB} \\
\hline
{\bf SVM} &87.1\% &92\%\\
\hline
{\bf Naive Bayes}  &88.4\% &89.6\% \\
\hline
\end{tabular}	
\caption{Average classification accuracies. Both classifiers show improvement in accuracies with features generated from our KB.}
\label{tab:classifier}
\end{table}

\section{Conclusions and Future Work}
\label{sec:conc}

In this paper, we described the construction of TeKnowbase, a knowledge-base of technical concepts related to computer science. Our approach consisted of two steps -- constructing a dictionary of terms related to computer science and to extract relationships among them. We made use of both structured and unstructured information source to extract relationships. Our experiments showed an accuracy of about 88\% on a subset of triples. We further used our KB in a classification task and showed how the features generated using the KB can increase classification accuracy.

There are lot of improvements that can be made to our system, purely to increase coverage. We used simple techniques, such as surface patterns, to extract relationships from textual sources. We can try more complex, supervised techniques to do the same. In order to extract unknown relationships, we are interested in exploring open IE techniques in more detail, particularly in identifying interesting and uninteresting relationships.


\end{document}